\documentclass{article}

\usepackage[english]{babel}

\usepackage[letterpaper,top=2cm,bottom=2cm,left=3cm,right=3cm,marginparwidth=1.75cm]{geometry}

\usepackage{subcaption}
\usepackage{amsmath}
\usepackage{graphicx}
\usepackage[colorlinks=true, allcolors=blue]{hyperref}

\title{Contrast Sensitivity in Multimodal Large Language Models: \\A Psychophysics-Inspired Evaluation}

\author{Pablo Hernández-Cámara$^{a,*}$, Alexandra Gomez-Villa$^{b,c}$, Jose Manuel Jaén-Lorites$^{d}$,\\ Jorge Vila-Tomás$^{a}$, Valero Laparra$^{a}$, Jesús Malo$^{a}$\\ \\$^{a}$ Image Processing Lab, Universitat de València, Spain \\ $^{b}$ Computer Vision Center, Spain \\ $^{c}$ Universitat Autònoma de Barcelona, Spain \\ $^{d}$ Center for Biomaterials and Tissue Engineering, Universitat Politecnica de Valencia, Spain \\ $^{*}$ Corresponding author: pablo.hernandez-camara@uv.es}

\begin{document}
\maketitle

\begin{abstract}
Understanding how Multimodal Large Language Models (MLLMs) process low-level visual features is critical for evaluating their perceptual abilities and has not been systematically characterized. Inspired by human psychophysics, we introduce a behavioural method for estimating the Contrast Sensitivity Function (CSF) in MLLMs by treating them as end-to-end observers. Models are queried with structured prompts while viewing noise-based stimuli filtered at specific spatial frequencies. Psychometric functions are derived from the binary verbal responses, and contrast thresholds (and CSFs) are obtained without relying on internal activations or classifier-based proxies. Our results reveal that some models resemble human CSFs in shape or scale, but none capture both. We also find that CSF estimates are highly sensitive to prompt phrasing, indicating limited linguistic robustness. Finally, we show that CSFs predict model performance under frequency-filtered and adversarial conditions. These findings highlight systematic differences in frequency tuning across MLLMs and establish CSF estimation as a scalable diagnostic tool for multimodal perception.
\end{abstract}

\section{Introduction}

\begin{figure}[ht]
    %\vspace{-0.5cm}
    \centering
    \includegraphics[width=\textwidth]{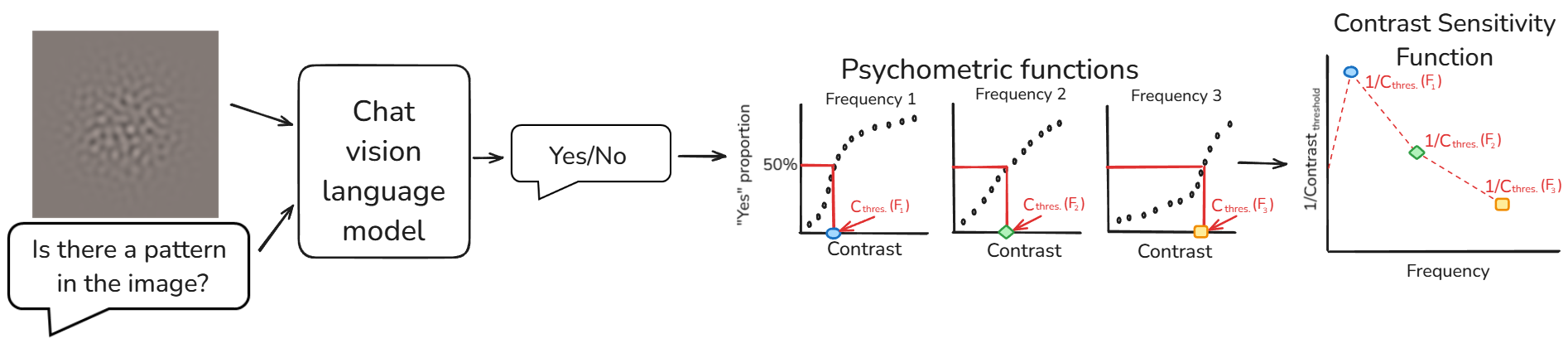}
        \vspace{-0.5cm}
    \caption{\textbf{Method summary:} Overview of the proposed psychophysics-inspired method for estimating contrast sensitivity functions (CSFs) in multimodal large language models (MLLMs). Models are presented with bandpass-filtered noise images at varying spatial frequencies and contrast levels, accompanied by natural language prompts. The binary (“yes/no”) responses are used to build psychometric functions, from which contrast detection thresholds are extracted. CSFs are computed as the inverse of these thresholds and analyzed across models and prompts.}
    \label{fig_method}
\end{figure}

The ability of a visual system to perceive luminance modulations (i.e., contrast) across different spatial frequencies is fundamental to its performance~\cite{goodman05}. In human vision, this ability is captured by the Contrast Sensitivity Function (CSF), which quantifies sensitivity to patterns as a function of contrast and spatial frequency~\cite{Campbell68}. The CSF determines how much detail we can perceive and is shaped by multiple factors, including temporal frequency~\cite{Kelly79}, luminance~\cite{Barten99,Wuerger20}, chromatic direction~\cite{Mullen85}, neural mechanisms~\cite{Derrington84}, eccentricity~\cite{Diez11}, and age~\cite{Owsley83}.

A familiar example is reading a billboard: as viewing distance increases, fine text becomes illegible because high spatial frequencies exceed human contrast sensitivity thresholds. This phenomenon highlights how our visual system is more sensitive to certain spatial frequencies while it cannot perceive fine details beyond a certain threshold, defining what the system can reliably perceive, ignore, or miss entirely.

Contrast sensitivity plays a critical role not only in biological vision but also in artificial perceptual systems. It defines the effective bandwidth over which models can extract relevant information from visual inputs, shaping how they filter signals, what kinds of detail they retain, and where their perceptual blind spots may lie. By analyzing the CSF of a system, we can predict its performance on new stimuli, identify its strengths and weaknesses, and assess its ability to generalize across different visual environments. This principle is essential in designing and evaluating artificial vision systems and is particularly relevant for multimodal systems, where visual perception must interact with downstream tasks such as language generation.

While previous studies have explored the CSF of convolutional nets \cite{li2022contrast, Akbarinia23,cai2025computer} and transformer-based vision models \cite{Akbarinia23, cai2025computer}, they have largely focused on unimodal vision architectures. Therefore, no work has yet explored Multimodal Large Language Models, i.e. multimodal systems capable of integrating vision and language through generative responses and maintaining a conversation. For instance, \cite{li2022contrast} showed that autoencoders with simpler architectures approximate the human CSF more closely. Later, \cite{Akbarinia23} extended this analysis to over 200 pre-trained convolutional neural networks (CNN's) across different vision tasks, revealing that human-like CSF's emerge more prominently in models optimized for low-level vision. More recently, \cite{cai2025computer} analyzed the vision part of some modern transformer-based foundational models and found that any of the analyzed models show a contrast sensitivity pattern similar to the human one, with many showing much lower sensitivity than humans at lower frequencies.

From the methodological point of view all these methods rely on arbitrary internal readout strategies to define the CSF. This introduces additional assumptions and complicates the interpretation of the results. Some methods use classification-based methods \cite{Akbarinia23}, where a classifier is trained to detect gratings at different contrasts, conflating model perception with the classifier’s performance. Others rely on Euclidean distance \cite{li2022contrast} or cosine similarity \cite{cai2025computer} in feature space, assuming that contrast perception corresponds to embedding differences, which imposes a potentially unnatural metric.

In this work, we propose a novel behavioural method for estimating the CSF of multimodal large language models (MLLMs) that mirrors human psychophysical experiments. Inspired by classical human psychophysics and instead of relying on internal representations, our method treats the model as an end-to-end observer: it is shown a stimulus image and asked, via natural language prompts, whether a pattern is present. For each fixed frequency, by varying contrast, we construct a psychometric function and derive the corresponding contrast detection threshold, just as in human experiments. This allows us to estimate the model’s CSF without any internal access, classifier training, or metric assumptions through a direct measurement of the contrast sensitivity of MLLMs. It allows us to assess how well these models capture visual information, understand where they may struggle and diagnose potential limitations, biases, and vulnerabilities. Importantly, our goal is not to enforce human-model alignment, but to provide a general, interpretable, and prompt-aware framework for probing low-level visual perception. Focusing on the human-model alignment, while not aiming to enforce it, we use the human CSF as a reference to shed light on its similarities and divergences.

Therefore, our contributions are:  (1) We introduce a novel, prompt-based methodology for estimating the contrast sensitivity function (CSF) in multimodal models, inspired by human psychophysics and free from internal readouts or custom metrics. (2) We demonstrate this approach by computing the CSF across a diverse set of open-source MLLMs, analyzing their sensitivity across spatial frequencies. (3) We assess the prompt robustness of each model's CSF, revealing the impact of linguistic variation on visual sensitivity estimation. (4) We compare model CSFs with human psychophysical data to contextualize frequency tuning patterns without assuming human-likeness as an optimization target. (5) We explore how the CSF can be used as a diagnostic tool to predict model performance on downstream tasks such as classification and adversarial robustness.

By bridging psychophysical testing and multimodal evaluation, our work offers a scalable and interpretable framework for studying visual perception in next-generation language models.

\section{Methods}
\label{sec:methods}

Our methodology builds on principles from classical human psychophysics, where contrast sensitivity is measured through behavioural testing. Rather than analyzing internal representations, we treat MLLMs as observers and probe their visual sensitivity using structured stimuli and carefully phrased prompts. 
This framework, illustrated in Figure~\ref{fig_method}, enables the estimation of contrast sensitivity functions (CSFs) in a way that is directly comparable to human perceptual testing, yet fully model-agnostic. It requires no access to internal activations, no classifier training, and no predefined similarity metrics.

The following subsections describe the design of the visual stimuli, the contrast sensitivity testing procedure, and the MLLMs evaluated in this study.

\subsection{Stimuli design}

To estimate the CSF of MLLMs, we generated a controlled set of visual stimuli inspired by classical human psychophysics~\cite{Kingdom16}. To generate the visual stimuli used for CSF estimation, we followed the method introduced by Cai et al.~\cite{cai2025computer}, ensuring consistency with recent work in vision modelling and allowing for direct comparability with human CSF estimations.

Following their design choices, we used bandpass-filtered noise images instead of sinusoidal gratings. Although both stimulus types are valid in psychophysics, bandpass-filtered noise introduces statistical variability while maintaining frequency specificity, improving robustness by avoiding the regularity and predictability of simple gratings \cite{cai2025computer, vilatomas25}.

Following prior work \cite{cai2025computer} and to ensure comparability with human psychophysical data, each stimulus was generated by applying a bandpass filter in the Fourier domain to two-dimensional Gaussian noise. Filters were centered on spatial frequencies logarithmically spaced between $0$ and $32$ cycles per degree (cpd), covering the range typically used in human CSF studies. We used a sampling resolution of $64$ pixels per degree, typical of modern displays. Stimuli were generated at a resolution of $256 \times 256$ pixels, corresponding to a visual field of $4 \times 4$ degrees, covering the human foveal region. Each image was normalized to a mean luminance of $40$ $cd/m^2$, ensuring that only contrast varied across conditions. Luminance values of these achromatic images were then converted to RGB format for compatibility with MLLMs requiring three-channel input.

For each spatial frequency, we generated images at multiple contrast levels. $160$ contrast values were linearly spaced between $0$ (no modulation) and $0.8$, covering the typical threshold region in CSF estimation. To account for stimulus variability, we created 25 statistically independent samples for each frequency–contrast pair by using different random noise seeds. This sampling strategy improves the reliability of the estimated psychometric functions by minimizing stimulus-specific artifacts.

\subsection{Task design: Prompt variability}

Following the paradigm of human psychophysics, we designed a task in which MLLMs are treated as experimental participants. In each trial, the model is presented with a stimulus image and a natural language prompt asking whether a visual pattern is present. This setup mirrors traditional human contrast sensitivity experiments, where participants report whether they detect a pattern at a given contrast level.

Because MLLMs are sensitive to prompt phrasing \cite{gavrikov2024vision, errica2024did}, we systematically varied the prompts used during testing. Following the methodology of Gavrikov et al.~\cite{gavrikov2024vision}, we constructed a prompt set comprising: 10 synonyms or variants of the noun "pattern" (e.g., structure, texture), 10 visibility-related adjectives or adverbs (e.g., visible, discernible), and 5 additional variations that modify word order and sentence structure. An example prompt is: “$<$image$>$ Is there a pattern in the image?”. All prompts followed the same binary decision format, enabling consistent comparison across linguistic variants. Each contrast–frequency–prompt combination was tested with 10 independently generated noise images, yielding robust psychometric functions for every spatial frequency and prompt formulation.

This design allows us to measure both frequency-specific sensitivity and prompt-induced variability, i.e. how stable its visual perception is under minor changes in prompt wording. Such analysis is particularly relevant in multimodal systems, where visual and linguistic modalities interact non-trivially. The full list of prompts is provided in~\ref{sec_appendix_a}.

\subsection{Psychometric function and contrast threshold}

To estimate the contrast sensitivity function (CSF) of each model, we adopted a procedure analogous to classical human psychophysical testing. For every combination of spatial frequency and prompt, the model was presented with multiple stimulus images at varying contrast levels and asked whether it could detect a pattern. For each contrast level, we computed the proportion of affirmative responses (i.e., yes answers), from which we constructed discrete psychometric functions (proportion of yes answers for each contrast and frequency).

We modelled these functions using a Weibull cumulative distribution, a standard choice in psychophysics for its flexibility in capturing detection thresholds and slopes:

\begin{equation}
    P(c) = 1 - e^{-(c/\alpha)^\beta}
\end{equation}

\noindent where $P(c)$ denotes the probability of detection at contrast $c$, and $\alpha$ and $\beta$ are the scale and shape parameters, respectively. This function was fitted independently for each spatial frequency and prompt, using data from 10 repeated stimulus presentations per condition.

The contrast detection threshold was then defined as the contrast level corresponding to a 50\% detection probability. These thresholds reflects the minimum contrast required for the model to reliably detect the presence of a pattern at a given spatial frequency. From this, we computed the contrast sensitivity in the standard way~\cite{Campbell68} as the inverse of the threshold:

\begin{equation}
    CSF(f) = \frac{1}{C_{threshold}(f)}
\end{equation}

\begin{figure}[ht]
    %\vspace{-0.5cm}
    \centering
    \includegraphics[width=\textwidth]{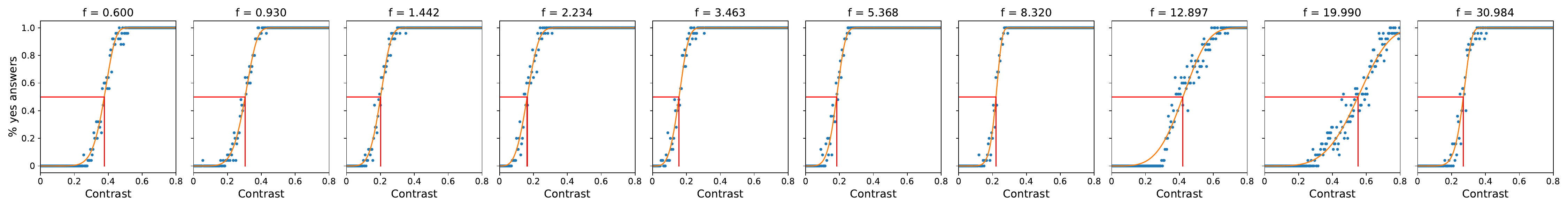}
        \vspace{-0.5cm}
    \caption{\textbf{Psychometric functions fitted:} Example of the psychometric functions fitted with the model's experimental data. Blue points correspond with the percentage of affirmative model answers for each contrast. The orange line is the fitted psychometric function, independent for each frequency. Finally, the red lines mark the 50\% detection point and the associated contrast threshold. In this case, the results are from the model Qwen2.5VL-7B with the prompt “$<$image$>$ Is there an observable pattern in the image? Respond just yes or no”.}
    \label{fig_psychometric_ejemplo}
\end{figure}

This yielded a contrast sensitivity function for each model, analogous to human CSFs in both shape and scale. Figure \ref{fig_psychometric_ejemplo} shows an example of the psychometric functions fitted to the Qwen2.5VL-7B experimental data for one of the considered prompts. We then averaged across all prompt variants to obtain a model-level CSF and analyzed prompt-specific functions separately to evaluate prompt robustness and variability in perceptual estimates.

\subsection{Models}
\label{models}

While our primary objective is not to benchmark all available models, we aim to demonstrate the applicability of our proposed psychophysics-inspired method for estimating contrast sensitivity functions (CSFs) in MLLMs. To this end, we evaluated a representative set of models to test our methodology. Specifically, we tested a range of open-source MLLMs across the 3B and 7B parameter scales, as well as a proprietary model included for illustrative comparison. All models were evaluated using the same set of contrast-varying stimuli and the same full set of 25 prompt formulations, with default inference settings. This controlled setup ensures that differences in CSF estimates are attributable to the models themselves rather than variations in experimental protocol.

The MLLMs included in this study are, for the 3B scale Qwen2.5VL-3B \cite{bai2025qwen2}, InternVL2.5-4B \cite{chen2024expanding} and InternVL3-2B \cite{zhu2025internvl3}; and for the 7B scale Qwen2.5VL-7B \cite{bai2025qwen2}, LLaVA-1.5-7B \cite{liu2023visual}, Magma-8B, \cite{yang2025magma}, InternVL2.5-8B \cite{chen2024expanding} and InternVL3-8B \cite{zhu2025internvl3}.

Although our primary focus is on MLLMs, our method can also be adapted to contrastive vision–language models such as CLIP \cite{radford2021learning} and SigLIP2 \cite{tschannen2025siglip}. We provide details and results in~\ref{sec_appendix_b} to demonstrate the generality of the framework.

\section{Results}
\label{sec:results}

In this section, we present the results of applying our psychophysics-inspired method to estimate contrast sensitivity functions (CSFs) across a range of multimodal large language models (MLLMs). We first visualize and compare the average CSF curves obtained from each model, highlighting distinct sensitivity profiles across spatial frequencies. We then examine the stability of these CSFs under prompt variation, revealing the extent to which linguistic phrasing affects visual sensitivity estimates.

Additional analyses, such as comparison with human CSFs and implications for downstream performance, are discussed in Sections~\ref{sec_human_alignment} and \ref{sec_applications}, where we explore how this method can serve as a tool for perceptual alignment and model diagnostics.

\begin{figure}[ht]
    \centering
    \includegraphics[width=0.90\textwidth]{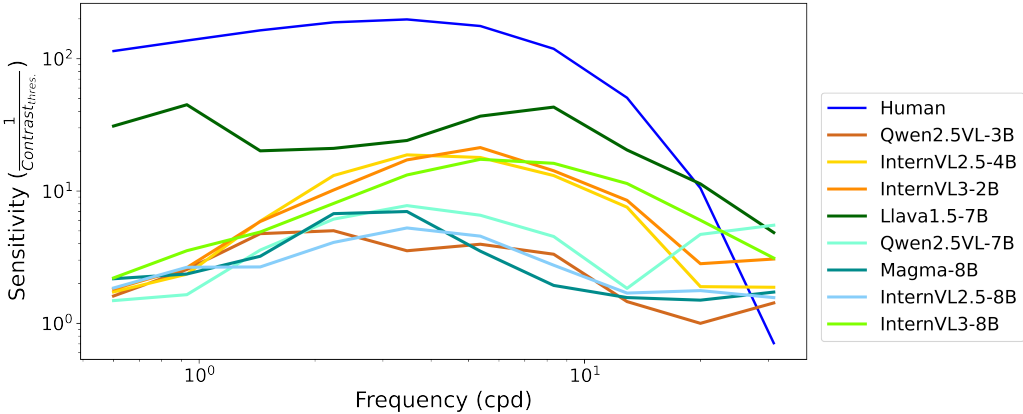}
    \caption{\textbf{Average contrast sensitivity functions (CSFs) for all tested models:} Each curve represents the average CSF across 25 prompt variants per spatial frequency. Note that we used log-log scales for better visualization. The human CSF (of the Standard Spatial Observer~\cite{Watson02}) is included for reference as the dark blue line.}
    \label{fig_csf_results}
\end{figure}

\subsection{Model's CSFs}

Figure~\ref{fig_csf_results} shows the average contrast sensitivity function (CSF) for each model, computed by aggregating responses across all prompt variants per spatial frequency. For comparison, the human CSF is included as a reference (blue line), exhibiting the classic bandpass shape with peak sensitivity around 4–6 cycles per degree (cpd). These CSF curves illustrate how each model responds to spatial frequency structure in the visual input and reveal key differences between them.

Across models, we observe substantial variability in both the shape and scale of the CSFs. Some models produce curves with a distinguishable peak at intermediate frequencies, while others display flatter or monotonic profiles. This suggests that some models exhibit frequency selectivity, while others do not.

In terms of magnitude, most models either overestimate or underestimate sensitivity in specific frequency bands. Notably, sensitivity is mainly underestimated at low frequencies, where humans are most sensitive, and some models overestimate at higher frequencies, where human performance typically declines. These deviations highlight important differences in how MLLMs process low-level spatial structure.

Despite these differences, the method consistently produces interpretable and discriminable CSFs across models. This demonstrates its utility as a general-purpose diagnostic tool, but also raises the question of how stable these estimates are under variations in natural language phrasing, i.e. prompt variability, a key factor in multimodal systems.

\subsection{Prompt Dependence}

Next, we analyze how prompt phrasing affects the reliability of CSF estimates. Despite using fixed visual stimuli, MLLMs can exhibit variable behaviour depending on how the image is queried. Prior work has shown that even minor changes in prompt phrasing can lead to significantly different outputs in LLMs, affecting not only their textual generations but also their factual, semantic, and classification behaviour \cite{errica2024did}.

We quantified prompt robustness using entropy and consistency metrics from Errica et al. \cite{errica2024did}. Specifically, we measure how robust each model's responses are to the 25 prompt formulations used in the CSF estimation procedure. We averaged both metrics across all frequency-contrast conditions to obtain model-level scores. These measures offer a complementary view of robustness: entropy captures how often model decisions fluctuate across prompts, while consistency reflects the internal coherence of model behaviour.

\begin{figure}[ht]
    \centering
    \includegraphics[width=0.48\textwidth]{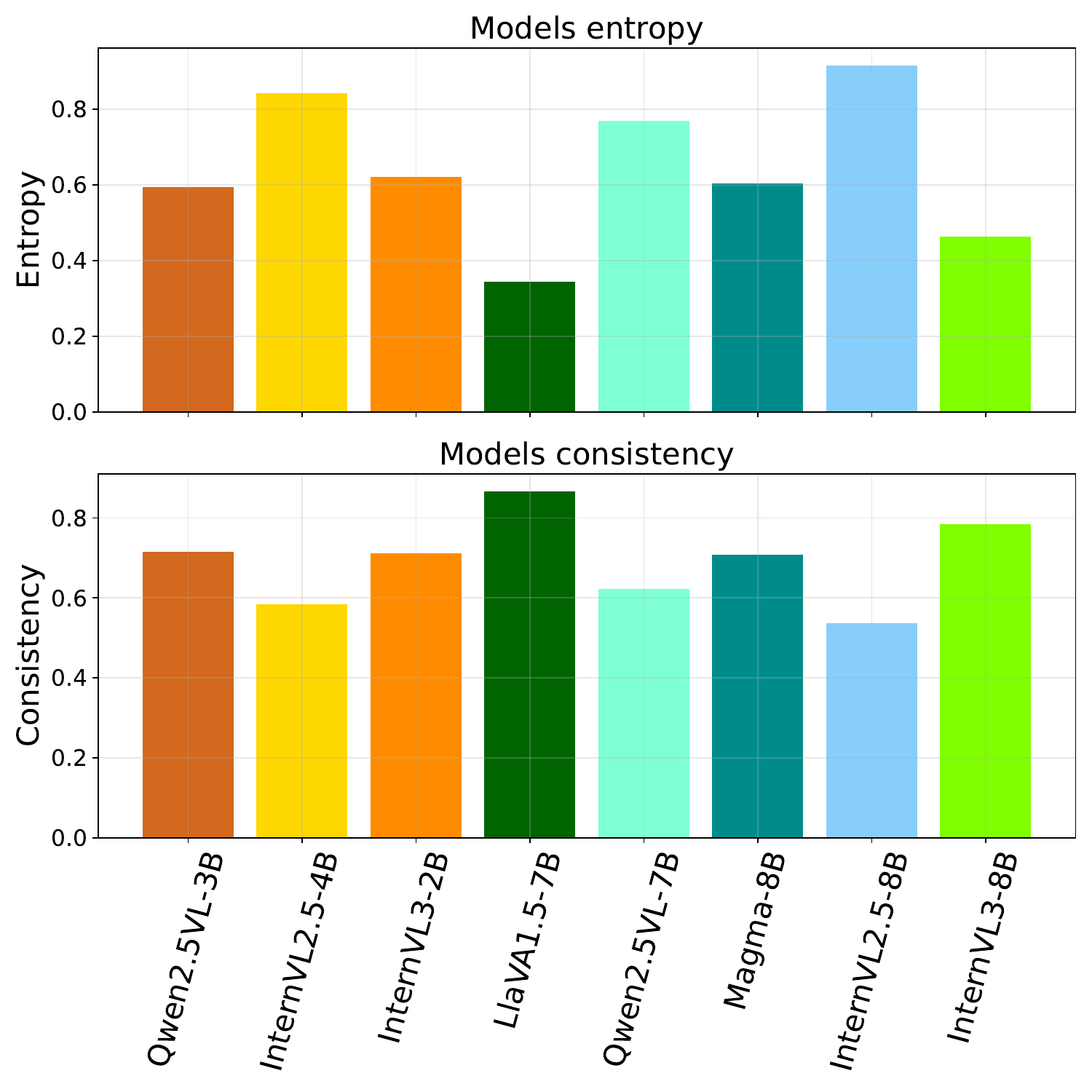}
    \vspace{-0.15cm}
    \caption{\textbf{Prompt sensitivity and consistency across models} Left: average entropy of model responses across 25 prompt formulations, indicating how much a model’s detection behaviour varies with phrasing (higher values = less stable). Right: prompt consistency, computed as the mean pairwise agreement between prompt responses per contrast–frequency condition (higher values = more stable). These metrics quantify how robust each model’s CSF estimation is to natural language variation in the query.}
    \vspace{-0.1cm}
\label{fig_prompt_dependence_metrics}
\end{figure}

Figure~\ref{fig_prompt_dependence_metrics} summarizes these metrics across models. Each bar represents the average sensitivity (top panel) or consistency (bottom panel) of a model’s responses across all frequency–contrast pairs. Again, we observe substantial differences across models: some models, such as the Intern2.5 family, exhibit high entropy and low consistency, indicating strong prompt dependence, i.e. small changes in phrasing lead to significantly different outputs. In contrast, Llava1.5-7B shows lower entropy and higher consistency, suggesting more stable responses across linguistic formulations. Overall, these results highlight that even when visual stimuli are held constant, the estimated contrast sensitivity is influenced by linguistic input, emphasizing that prompt formulation is a key source of variability in multimodal perceptual probing. While prompt phrasing introduces measurable variance in estimated CSFs, an equally important question is how these CSFs compare structurally to human vision.

\section{Human Alignment}
\label{sec_human_alignment}

To explore how model contrast sensitivity aligns to human vision, we compared the model CSFs to the Standard Spatial Observer human CSF \cite{Watson02} using two metrics: The Pearson correlation ($\rho_{Pearson}$), to assess similarity in shape; and the Root Mean Square Error (RMSE) to assess differences in absolute deviation, quantifying both shape and scale differences between the model and human CSFs. Note that $RMSE = 0$ implies $\rho = 1$, but high correlation does not imply low RMSE (a scaled version of the human CSF would show $\rho \approx 1$ but high RMSE). This distinction allows us to interpret high correlation as an indicator of structural similarity (e.g., bandpass behaviour), while RMSE reflects how closely the overall contrast sensitivity levels align.
Our goal is not to rank models by perceptual fidelity, but to characterize how their CSFs structurally compare with a well-characterised biological system, the human CSF.

\begin{table}[h]
    \centering
    \begin{tabular}{l||c|c}
    Model & $\rho_{Pearson}$ $\uparrow$ & RMSE $\downarrow$ \\\hline\hline
    %\textbf{Propietary} & & \\
    %Gemini 2.0-flash & 0.46 & 78.5 \\\hline
    \textbf{3B Scale} & & \\
    %Qwen 2.5VL (3B) & \textbf{0.83} & 103.5 \\
    Qwen 2.5VL (3B) & \textbf{0.86} & 131.6 \\
    %Blip 2 (2.7B) & 0.20 & 90.8 \\
    %Intern 2.5 (4B) & 0.02 & 121.4  \\\hline
    InternVL 2.5 (4B) & 0.69 & 125.2  \\
    InternVL 3 (2B) & 0.59 & 125.3  \\\hline
    \textbf{7B Scale} & & \\
    %InstructBlip-Vicuna (7B) & -0.46 & 106.0 \\
    %Llava 1.5 (7B) & 0.70 & \textbf{50.4} \\
    Llava 1.5 (7B) & 0.54 & \textbf{109.8} \\
    %Qwen 2.5VL (7B) & -0.23 & 129.7 \\
    Qwen 2.5VL (7B) & 0.31 & 130.6 \\
    %Magma (8B) & 0.24 & 111.4 \\
    Magma (8B) & 0.76 & 131.2 \\
    %Intern 2.5 (8B) & 0.00 & 128.9 \\\hline
    InternVL 2.5 (8B) & 0.84 & 131.6 \\
    InternVL 3 (8B) & 0.37 & 126.4 \\\hline
    %\textbf{Contrastive} & & \\
    %CLIP-L14 & -0.49 & 111.2 \\
    %CLIP-b16-224 & 0.46 & 106.1 \\
    %SigLIP-SO400M-P14 & -0.71 & 133.8\\
    %SigLIP-b16-224 & 0.45 & \textbf{102.2}\\
    \end{tabular}
    \caption{\textbf{Comparison between each model’s average CSF and the human CSF:} Pearson correlation ($\rho_{Pearson}$) measures shape similarity; root mean squared error (RMSE) measures deviation in absolute sensitivity. These values highlight perceptual differences across models, with no expectation of human matching. Highest correlation and lowest error values are highlighted in bold to illustrate models that most closely resemble human CSF shape and scale.}
    \label{tab_human_alignment}
\end{table}

Table~\ref{tab_human_alignment} presents these alignment results. All the models show positive correlation with human CSF; however, some models show better shape alignment than others (e.g., Qwen2.5VL-3B with $\rho_{Pearson} = 0.86$), while others better match overall sensitivity magnitude (e.g., Llava1.5-7B with $RMSE = 109.8$). In general, any of the analyzed models match human contrast sensitivity at a quantitative level; however, some of the models partially reproduce its qualitative shape.

These findings illustrate the diversity in frequency tuning across high-performing MLLMs and demonstrate how the CSF can serve as a tool to interpret a model’s perceptual filtering behaviour, even when alignment with human sensitivity is not the goal. This comparison is not intended as a benchmark but as a tool to situate models within an interpretable perceptual reference frame.

\section{Applications}
\label{sec_applications}

While our primary goal is to introduce a method for estimating contrast sensitivity functions in multimodal large language models, the CSF itself serves as a rich diagnostic tool. Once estimated, it provides insight into a model’s perceptual behaviour and can be applied to a variety of downstream analyses.

In this section, we demonstrate how the CSF can be used to analyze model classification robustness under two complementary perturbation scenarios: (1) frequency-specific removal, where information at particular spatial frequencies is suppressed, and (2) frequency-specific adversarial noise, where perturbations are added at targeted frequencies. Critically, both manipulations are controlled to produce constant image distortion, i.e. constant RMSE with original images, across all frequency bands, ensuring that differences in classification accuracy reflect frequency-specific vulnerabilities.

\begin{figure}[ht]
    \centering

\begin{subfigure}{0.98\textwidth}
    \centering
    \includegraphics[width=0.9\textwidth]{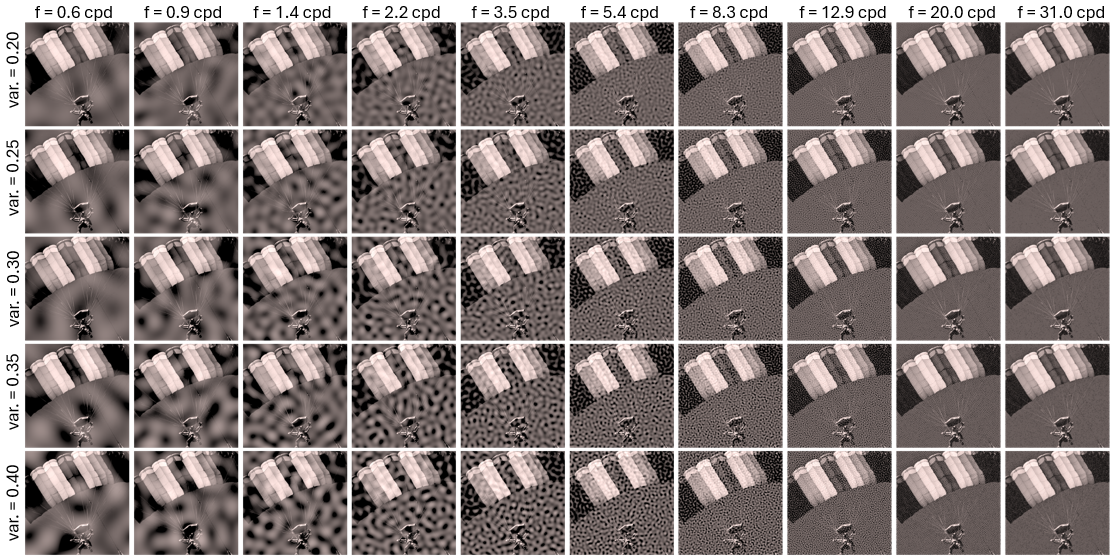}
    \caption{Frequency-specific adversarial images.}
    \label{ejemplo_adversarial}
\end{subfigure}
\vfill
\begin{subfigure}{0.98\textwidth}
    \centering
    \includegraphics[width=0.9\textwidth]{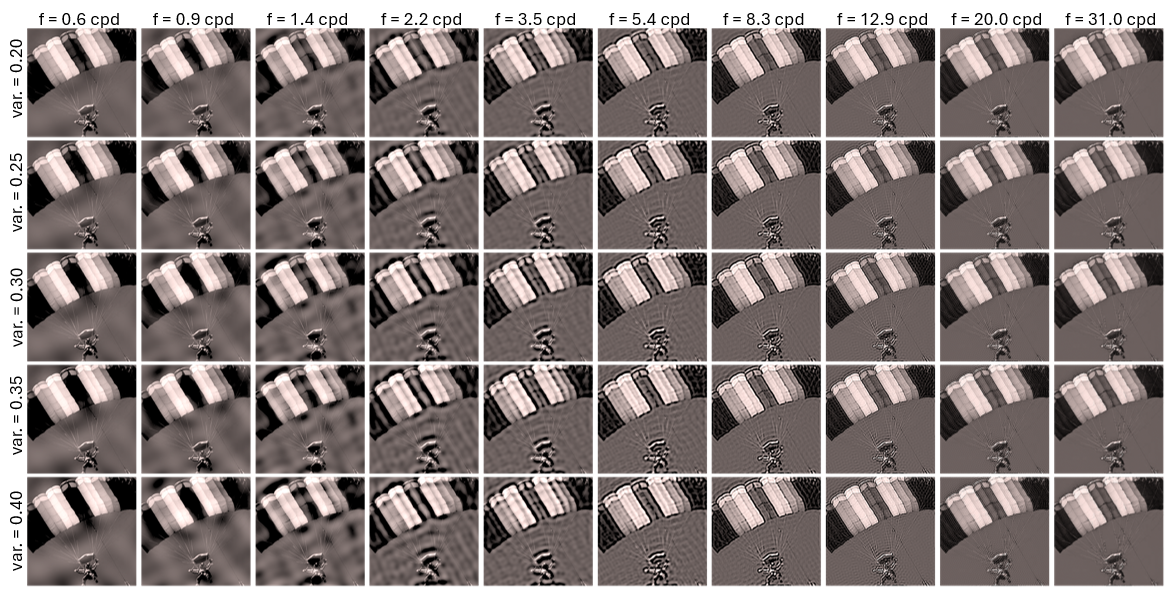}
    \caption{Frequency-specific filtered images.}
    \label{ejemplo_filter}
\end{subfigure}

    \caption{\textbf{Frequency-specific adversarial and filtered images:} Example of the adversarial (top) and filtered (bottom) effect on an image depending on the filtered frequency band (columns) and the induced variance (rows).}
    %\label{ejemplo_filter}
\end{figure}

\subsection{CSF and Robustness to Frequency Filtering}

We first investigated whether a model's contrast sensitivity function predicts its robustness when specific spatial frequency information is removed from input images. For each spatial frequency band, we applied a bandpass suppression filter in the Fourier domain that attenuated the corresponding frequency range while leaving other frequencies intact.

To ensure fair comparison across frequency bands, we calibrated the suppression strength for each filter such that all filtered images exhibited the same variance reduction and consequently, the same RMSE relative to the original unfiltered images. This normalization is critical: it guarantees that all perturbations introduce equal amounts of distortion in pixel space, so any differences in classification accuracy can be attributed to the removal of specific frequency content rather than to differences in overall image degradation. Figure \ref{ejemplo_filter} shows the resulting images depending on the frequency and the introduced variance.

We then evaluated each model's classification accuracy on a standard image dataset, Imagenette \cite{Howard_Imagenette_2019}, using both original and frequency-filtered images. For each frequency band, we computed the relative accuracy drop as the difference in accuracy between the unfiltered and the filtered images for each frequency and variance.

The hypothesis is that models should be most vulnerable to removing frequencies where they are most sensitive, as indicated by their CSF. In other words, suppressing frequency bands where the CSF peaks should cause larger accuracy drops than suppressing bands where sensitivity is low.

\begin{figure}[ht]
    \centering
    \includegraphics[width=0.96\textwidth]{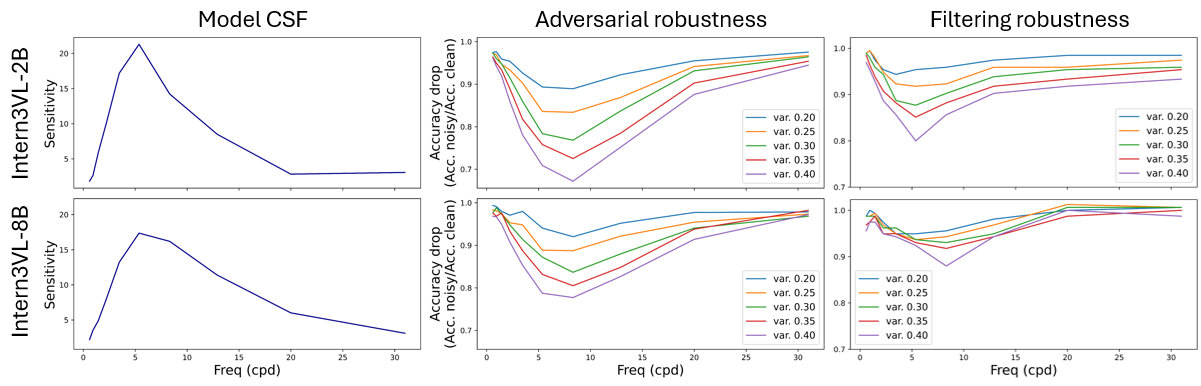}
    \caption{\textbf{Relationship between model contrast sensitivity (left) and classification accuracy degradation under different frequency perturbations (center, right) for Intern VL3-2B (top) and Intern VL3-8B (bottom):} The contrast sensitivity function (CSF) reflects each model’s estimated sensitivity across spatial frequencies. The middle column shows the relative change in classification accuracy when noise of different frequencies is added to the input images, and the right column shows the relative change in accuracy when information from specific frequencies is removed through filtering. Accuracy values are reported relative to performance on unaltered images. Both models exhibit a clear inverse relationship between contrast sensitivity and robustness: the largest accuracy reductions occur at frequencies where the model shows peak sensitivity.}
    \label{fig_csf_vs_accuracy}
\end{figure}

Figure \ref{fig_csf_vs_accuracy} shows the CSF (left) and relative classification accuracy drop (right) for two representative models of the same family that show a band-pass behaviour: Intern3VL-2B and Intern3VL-8B. In both cases, we observe a clear inverse relationship: the largest accuracy reductions occur at frequencies where the model exhibits peak sensitivity according to its CSF. Also, note how the more sensitive model suffers a higher accuracy reduction.

This pattern suggests that the CSF captures not just passive sensitivity but functionally critical frequency bands. Models rely most heavily on the frequencies they can detect best, making these bands both their perceptual strength and a potential vulnerability when that information is removed.

\subsection{CSF and Robustness to Frequency Adversarial Test}

In addition to frequency-specific removal, we evaluated how classification accuracy is affected by adding frequency-specific perturbations to input images. This complements the filtering experiment by testing vulnerability to added information rather than removed information.

For each spatial frequency band, we generated bandpass-filtered Gaussian noise centered on that frequency. Again, we calibrated the noise amplitude across all frequency bands to ensure that the noise had constant variance regardless of frequency, and therefore the resulting perturbed images exhibited constant RMSE relative to the original unperturbed images. This equalization ensures that all perturbations introduce equal amounts of pixel-level distortion, allowing us to isolate the effect of frequency content on model robustness. Figure \ref{ejemplo_adversarial} shows the resulting images depending on the frequency and the introduced variance. Note how, for the same variances, the adversarial images are more noticeable than the corresponding filtered images. This is because in the adversarial scenario, the added noise is stationary, i.e. flat, and not related to the images.

The hypothesis mirrors the filtering case: models should be most vulnerable to adversarial noise at frequencies where they are most sensitive. Adding noise at CSF peak frequencies should degrade accuracy more than adding equal-magnitude noise at frequencies where sensitivity is low.

Figure~\ref{fig_csf_vs_accuracy} shows the CSF (left) and classification accuracy degradation (center) for the same Intern VL 3-2B and Intern VL 3-8B models under frequency-localized adversarial noise. Again, we observe a strong correspondence: the largest accuracy drops occur at spatial frequencies where the model exhibits peak sensitivity. For both models, adversarial noise at CSF peak frequencies causes substantially greater performance degradation than noise at frequencies where sensitivity is lower, despite all noise perturbations having identical RMSE. It is interesting to highlight how the models are more affected by this adversarial noise than in the previous filtered scenario, similarly to human perception.

This finding reveals a fundamental asymmetry: while models are tuned to detect patterns at their peak sensitivity frequencies, this same tuning makes them disproportionately vulnerable to adversarial manipulation at those frequencies. The CSF thus identifies not only perceptual capabilities but also perceptual attack surfaces.

\vspace{0.5cm}

Together, these experiments demonstrate that a model's CSF is both descriptive and predictive. It describes the frequency tuning of the visual system and predicts task-relevant vulnerabilities under frequency-specific manipulations. Importantly, these vulnerabilities emerge even when total image distortion RMSE is held constant, indicating that the CSF captures frequency-specific dependencies that are not apparent from pixel-level metrics alone.
These findings suggest that CSF profiling could serve as a diagnostic tool for identifying structural weaknesses in multimodal vision systems, with potential implications for improving robustness through frequency-aware training strategies or input preprocessing.

\section{Conclusions}
\label{sec_conclusion}

Here, we introduced a psychophysics-inspired methodology for estimating contrast sensitivity functions in multimodal large language models. 

Unlike prior approaches using internal activations or trained classifiers, we treat models as end-to-end observers, mirroring human perceptual experiments, by querying them with structured stimuli and natural language prompts. This allows us to measure how model sensitivity varies across spatial frequencies in a direct, interpretable way.

We applied our method to a diverse set of open-source MLLMs and showed that the resulting CSFs capture meaningful differences in frequency tuning across models. Some exhibit clear bandpass sensitivity similar in shape to the human CSF, while others display flattened or irregular profiles. Beyond descriptive analysis, we demonstrated that CSFs correlate with model behaviour in downstream tasks. In particular, models tend to perform worse when visual information is removed or perturbed at the frequencies where they are most sensitive, highlighting a perceptual vulnerability aligned with their contrast tuning.

We also analyzed the influence of prompt phrasing, which introduces an additional source of variability in CSF estimation. Using entropy- and consistency-based metrics inspired by recent work on LLMs, we found that some models are highly sensitive to minor changes in linguistic formulation, while others remain stable across rephrasings. This underscores the importance of accounting for language-induced variance when probing the perceptual behaviour of multimodal systems.

\subsection{Limitations and Future Work}

While our results validate the effectiveness and flexibility of the proposed methodology, some limitations remain and point to promising directions for future work.

First, although we aim to propose a general methodology for estimating CSFs in MLLMs, the present study focuses on a small, limited set of open-source models. Broader comparisons involving more architectures, training objectives, and parameter scales are left for future work. Large-scale benchmarking remains important future work, but it is beyond the scope of this paper due to its high computational cost. However, it will be critical to fully map the perceptual landscape of current and future models.

Another limitation concerns the use of synthetic stimuli. While bandpass-filtered noise or sinusoidal gratings are standard tools in human psychophysics due to their controllable frequency structure, they lack the rich spatial correlations found in natural images. Prior work has shown that both biological and artificial systems respond differently to naturalistic versus artificial stimuli \cite{felsen2005natural}. Future extensions of our method could incorporate natural image stimuli with controlled frequency biases to enhance ecological validity and bridge the gap between perceptual sensitivity and real-world behaviour.

In parallel, our approach opens the door to a growing line of work aimed at bridging cognitive neuroscience and deep learning through structured, hypothesis-driven evaluation. In particular, the recent MindSet Vision framework (both high-level~\cite{biscione2024mindset} and low-level~\cite{vilatomas25}) proposes a suite of visual phenomena from cognitive psychology that could complement our CSF-based analysis. Integrating perceptual diagnostics like ours with controlled tests of visual illusions, invariance, and attention biases could lead to a more holistic understanding of how MLLMs process visual information.

More broadly, while benchmarks such as Visual Arena \cite{chiang2024chatbot} provide valuable high-level task evaluations, they often overlook low-level perceptual factors. Recent works \cite{biscione2024mindset, cai2025computer, vilatomas25} and our approach suggest that continuous psychophysical profiling may offer task-agnostic insights into model robustness, generalization, and visual grounding.

Finally, CSF characterization could support the attribution and detection of model-generated content. As synthetic images become increasingly realistic, recent work \cite{you2025images} has shown that even imperceptible generative artifacts can be detected by classifiers—suggesting that frequency-sensitive signatures may remain latent in model outputs. CSF profiling could thus provide a scalable, interpretable tool for watermarking, provenance, and content auditing in future generative systems.

\vspace{0.5cm}

Overall, we believe that contrast sensitivity and our framework provide a promising new lens for evaluating and interpreting the perceptual behaviours of emerging multimodal systems. It opens the door to prompt-aware, frequency-sensitive evaluation protocols and encourages a deeper integration of principles from visual neuroscience into the evaluation of artificial perceptual systems.

\section*{Acknowledgments}

This work was supported in part by MICIIN/FEDER/UE under Grants PID2020118071GB-I00, PDC2021-121522-C21 (funded by MCIN/AEI/10.13039/501100011033
and the EU NextGenerationEU/PRTR) and Grant PID2023-152133NB-I00; in part by
Spanish MIU under Grant FPU21/02256; in part by Generalitat Valenciana under
Projects GV/2021/074, CIPROM/2021/056 and CIAPOT/2021/9; and in part by the grant
BBVA Foundations of Science program: Maths, Stats, Comp. Sci. and AI (VIS4NN). The authors gratefully acknowledge the computer resources at Artemisa and the technical support provided by the European Union through the 2014-2020 ERDF Operative Programme of
Comunitat Valenciana, project IDIFEDER/2018/048.

\bibliographystyle{plain}
\bibliography{egbib}

\begin{thebibliography}{10}

\bibitem{dictionary_flat}
Dictionary.com. flat.
\newblock \url{https://www.thesaurus.com/browse/flat}, 2025.
\newblock Accessed: 05/09/2025.

\bibitem{dictionary_pattern}
Dictionary.com. pattern.
\newblock \url{https://www.thesaurus.com/browse/pattern}, 2025.
\newblock Accessed: 05/09/2025.

\bibitem{dictionary_visible}
Dictionary.com. visible.
\newblock \url{https://www.thesaurus.com/browse/visible}, 2025.
\newblock Accessed: 05/09/2025.

\bibitem{Akbarinia23}
Arash Akbarinia, Yaniv Morgenstern, and Karl~R. Gegenfurtner.
\newblock Contrast sensitivity function in deep networks.
\newblock {\em Neural Networks}, 164:228--244, 2023.

\bibitem{bai2025qwen2}
Shuai Bai, Keqin Chen, Xuejing Liu, Jialin Wang, Wenbin Ge, Sibo Song, Kai Dang, Peng Wang, Shijie Wang, Jun Tang, et~al.
\newblock Qwen2. 5-vl technical report.
\newblock {\em arXiv preprint arXiv:2502.13923}, 2025.

\bibitem{Barten99}
Peter G.~J. Barten.
\newblock Contrast sensitivity of the human eye and its effects on image quality, 1999.

\bibitem{biscione2024mindset}
Valerio Biscione, Dong Yin, Gaurav Malhotra, Marin Dujmovic, Milton~L Montero, Guillermo Puebla, Federico Adolfi, Rachel~F Heaton, John~E Hummel, Benjamin~D Evans, et~al.
\newblock Mindset: Vision. a toolbox for testing dnns on key psychological experiments.
\newblock {\em arXiv preprint arXiv:2404.05290}, 2024.

\bibitem{cai2025computer}
Yancheng Cai, Fei Yin, Dounia Hammou, and Rafal Mantiuk.
\newblock Do computer vision foundation models learn the low-level characteristics of the human visual system?
\newblock {\em CVPR}, 2025.

\bibitem{Campbell68}
F.~W. Campbell and J.~G. Robson.
\newblock Application of fourier analysis to the visibility of gratings.
\newblock {\em J. Physiol.}, 197(3):551--566, 1968.

\bibitem{chen2024expanding}
Zhe Chen, Weiyun Wang, Yue Cao, Yangzhou Liu, Zhangwei Gao, Erfei Cui, Jinguo Zhu, Shenglong Ye, Hao Tian, Zhaoyang Liu, et~al.
\newblock Expanding performance boundaries of open-source multimodal models with model, data, and test-time scaling.
\newblock {\em arXiv preprint arXiv:2412.05271}, 2024.

\bibitem{chiang2024chatbot}
Wei-Lin Chiang, Lianmin Zheng, Ying Sheng, Anastasios~Nikolas Angelopoulos, Tianle Li, Dacheng Li, Banghua Zhu, Hao Zhang, Michael Jordan, Joseph~E Gonzalez, et~al.
\newblock Chatbot arena: An open platform for evaluating llms by human preference.
\newblock In {\em Forty-first International Conference on Machine Learning}, 2024.

\bibitem{Derrington84}
A~M Derrington and P~Lennie.
\newblock Spatial and temporal contrast sensitivities of neurones in lateral geniculate nucleus of macaque.
\newblock {\em J. Physiol.}, 357(1):219--240, 1984.

\bibitem{Diez11}
M.A. Díez-Ajenjo, P.~Capilla, and M.J. Luque.
\newblock Red-green vs. blue-yellow spatio-temporal contrast sensitivity across the visual field.
\newblock {\em Journal of Modern Optics}, 58(19-20):1736--1748, 2011.

\bibitem{errica2024did}
Federico Errica, Giuseppe Siracusano, Davide Sanvito, and Roberto Bifulco.
\newblock What did i do wrong? quantifying llms' sensitivity and consistency to prompt engineering.
\newblock {\em arXiv preprint arXiv:2406.12334}, 2024.

\bibitem{felsen2005natural}
Gidon Felsen and Yang Dan.
\newblock A natural approach to studying vision.
\newblock {\em Nature neuroscience}, 8(12):1643--1646, 2005.

\bibitem{gavrikov2024vision}
Paul Gavrikov, Jovita Lukasik, Steffen Jung, Robert Geirhos, Bianca Lamm, Muhammad~Jehanzeb Mirza, Margret Keuper, and Janis Keuper.
\newblock Are vision language models texture or shape biased and can we steer them?
\newblock {\em ICLR}, 2025.

\bibitem{goodman05}
Joseph~W Goodman.
\newblock {\em Introduction to Fourier optics, 3rd ed.}
\newblock Roberts \& Co. Publishers, Englewood, CO, USA, 2005.

\bibitem{Howard_Imagenette_2019}
Jeremy Howard.
\newblock Imagenette: A smaller subset of 10 easily classified classes from imagenet, March 2019.

\bibitem{Kelly79}
D.~H. Kelly.
\newblock Motion and vision. ii. stabilized spatio-temporal threshold surface.
\newblock {\em J. Opt. Soc. Am.}, 69(10):1340--1349, 1979.

\bibitem{Kingdom16}
Frederick~A.A. Kingdom and Nicolaas Prins.
\newblock Chapter 4 - psychometric functions.
\newblock In Frederick~A.A. Kingdom and Nicolaas Prins, editors, {\em Psychophysics (Second Edition)}, pages 55--117. Academic Press, San Diego, USA, second edition edition, 2016.

\bibitem{li2022contrast}
Qiang Li, Alex Gomez-Villa, Marcelo Bertalmio, and Jesus Malo.
\newblock Contrast sensitivity functions in autoencoders.
\newblock {\em Journal of Vision}, 22(6):8--28, 2022.

\bibitem{liu2023visual}
Haotian Liu, Chunyuan Li, Qingyang Wu, and Yong~Jae Lee.
\newblock Visual instruction tuning.
\newblock {\em Advances in neural information processing systems}, 36:34892--34916, 2023.

\bibitem{Mullen85}
K~T Mullen.
\newblock The contrast sensitivity of human colour vision to red-green and blue-yellow chromatic gratings.
\newblock {\em The Journal of Physiology}, 359(1):381--400, 1985.

\bibitem{Owsley83}
Cynthia Owsley, Robert Sekuler, and Dennis Siemsen.
\newblock Contrast sensitivity throughout adulthood.
\newblock {\em Vis. Res.}, 23(7):689--699, 1983.

\bibitem{radford2021learning}
Alec Radford, Jong~Wook Kim, Chris Hallacy, Aditya Ramesh, Gabriel Goh, Sandhini Agarwal, Girish Sastry, Amanda Askell, Pamela Mishkin, Jack Clark, et~al.
\newblock Learning transferable visual models from natural language supervision.
\newblock In {\em International conference on machine learning}, pages 8748--8763. PMLR, 2021.

\bibitem{tschannen2025siglip}
Michael Tschannen, Alexey Gritsenko, Xiao Wang, Muhammad~Ferjad Naeem, Ibrahim Alabdulmohsin, Nikhil Parthasarathy, Talfan Evans, Lucas Beyer, Ye~Xia, Basil Mustafa, et~al.
\newblock Siglip 2: Multilingual vision-language encoders with improved semantic understanding, localization, and dense features.
\newblock {\em arXiv preprint arXiv:2502.14786}, 2025.

\bibitem{vilatomas25}
Jorge Vila-Tomás, Pablo Hernández-Cámara, Qiang Li, Valero Laparra, and Jesús Malo.
\newblock A turing test for artificial nets devoted to model human vision, 2025.

\bibitem{Watson02}
Andrew~B. Watson and Jesus Malo.
\newblock Video quality measures based on the {S}tandard {S}patial {O}bserver.
\newblock In {\em Proc. IEEE Int. Conf. Im. Proc. ICIP02}, pages 41--44, 2002.

\bibitem{Wuerger20}
S.~Wuerger, M.~Ashraf, M.~Kim, J.~Martinovic, M.~Pérez-Ortiz, and R.K. Mantiuk.
\newblock Spatio-chromatic contrast sensitivity under mesopic and photopic light levels.
\newblock {\em J. Vis.}, 20(4):23--23, 04 2020.

\bibitem{yang2025magma}
Jianwei Yang, Reuben Tan, Qianhui Wu, Ruijie Zheng, Baolin Peng, Yongyuan Liang, Yu~Gu, Mu~Cai, Seonghyeon Ye, Joel Jang, et~al.
\newblock Magma: A foundation model for multimodal ai agents.
\newblock In {\em Proceedings of the Computer Vision and Pattern Recognition Conference}, pages 14203--14214, 2025.

\bibitem{you2025images}
Zebin You, Xinyu Zhang, Hanzhong Guo, Jingdong Wang, and Chongxuan Li.
\newblock Are images indistinguishable to humans also indistinguishable to classifiers?
\newblock In {\em Proceedings of the Computer Vision and Pattern Recognition Conference}, pages 28790--28800, 2025.

\bibitem{zhu2025internvl3}
Jinguo Zhu, Weiyun Wang, Zhe Chen, Zhaoyang Liu, Shenglong Ye, Lixin Gu, Hao Tian, Yuchen Duan, Weijie Su, Jie Shao, et~al.
\newblock Internvl3: Exploring advanced training and test-time recipes for open-source multimodal models.
\newblock {\em arXiv preprint arXiv:2504.10479}, 2025.

\end{thebibliography}

\appendix
\section{Prompt design for MLLM}
\label{sec_appendix_a}

This is a list of the full set of prompts used to query the MLLMs during contrast sensitivity testing. Prompts were designed to vary in noun choice, adjective use, and sentence structure to evaluate prompt dependence. We select the "pattern" and "visible" synonyms from \emph{Thesaurus.com} \cite{dictionary_pattern, dictionary_visible}.

\begin{itemize}
    \item Pattern synonims:
    \begin{itemize}
        \item ``Is there a pattern on the image?"
        \item ``Is there an arrangement on the image?"
        \item ``Is there a spatial structure on the image?"
        \item ``Is there a design on the image?"
        \item ``Is there a noise on the image?"
        \item ``Is there a spatial variation on the image?"
        \item ``Is there a contrast modulation on the image?"
        \item ``Is there a composition on the image?"
        \item ``Is there a structure on the image?"
        \item ``Is there a motif on the image?"
    \end{itemize}
    \item Visible synonims:
    \begin{itemize}
        \item ``Is there a visible pattern on the image?"
        \item ``Is there a detectable pattern on the image?"
        \item ``Is there a discernible pattern on the image?"
        \item ``Is there a distinguishable pattern on the image?"
        \item ``Is there an evident pattern on the image?"
        \item ``Is there a noticeable pattern on the image?"
        \item ``Is there an observable pattern on the image?
        \item ``Is there a patent pattern on the image?
        \item ``Is there a perceptible pattern on the image?
        \item ``Is there an appreciable pattern on the image?
    \end{itemize}
    \item Word order:
    \begin{itemize}
        \item ``On the image, is there a pattern?"
        \item ``Is there on the image a pattern?"
        \item ``A pattern is there on the image?"
        \item ``Is a pattern there on the image?"
        \item ``There is a pattern on the image, is there?"
    \end{itemize}
\end{itemize}

\section{Extension to Contrastive Models}
\label{sec_appendix_b}

While our primary focus is on MLLMs with generative ability, our methodology can also be adapted to contrastive-trained multimodal models. These models do not support conversational prompting, but can still be evaluated as observers in our methodology, in this scenario, by computing similarities between image and text embeddings.

In this case, for each trial, we presented the model with an image (again, defined by a specific contrast and spatial frequency) and a pair of texts. We measured the cosine similarity between its image embedding and two candidate textual descriptions, such as "an image with a pattern" and "a flat image". The label with higher similarity was taken as the model's response. This allowed us to, similarly to MLLMs, construct psychometric functions and estimate contrast thresholds. To account for the linguistic variability, we also varied the positive and negative prompts as shown in section \ref{clip_prompts}. Particularly, the contrastive models included in this example are CLIP \cite{radford2021learning} and SigLIP2 \cite{tschannen2025siglip}.

\subsection{Contrastive-trained prompts}
\label{clip_prompts}

This is a list of the full set of prompts used to query the contrastive models during contrast sensitivity testing. Each combination includes two prompts, one positive reflecting the pattern present in the image and the other negative reflecting that there is no pattern. We use 20 different combinations. In the first ten, we leave untouched the positive one and vary the negative one with synonyms of "flat". In the last ten we leave untouched the negative one and vary the positive one with synonyms of "pattern".  We select the "pattern" and "flat" synonyms from \emph{Thesaurus.com} \cite{dictionary_pattern, dictionary_flat}.

\begin{itemize}
    \item Varying the negative prompt with synonyms of "flat":
    \begin{itemize}
        \item ``A flat image" - ``An image with a pattern"
        \item ``A monotonous image" - ``An image with a pattern"
        \item ``A plain image" - ``An image with a pattern"
        \item ``A uniform image" - ``An image with a pattern"
        \item ``A homogeneous image" - ``An image with a pattern"
        \item ``A constant image" - ``An image with a pattern"
        \item ``A unvarying image" - ``An image with a pattern"
        \item ``A unfluctuating image" - ``An image with a pattern"
        \item ``A even image" - ``An image with a pattern"
        \item ``A steady image", - ``An image with a pattern" 
    \end{itemize}
    \item Varying the positive prompt with synonyms of "pattern":
    \begin{itemize}
        \item ``A flat image" - ``An arrangement image"
        \item ``A flat image" - ``An image with spatial structure" 
        \item ``A flat image" - ``An image with a design"  
        \item ``A flat image" - ``A noisy image" 
        \item ``A flat image" - ``An image with spatial variations" 
        \item ``A flat image" - ``An image with contrast modulations"  
        \item ``A flat image" - ``A composite image"  
        \item ``A flat image" - ``A structured image"
        \item ``A flat image" - ``An image with a motif"  
        \item ``A flat image" - ``A non-flat image" 
    \end{itemize}
\end{itemize}

\subsection{Contrastive-trained CSF}
\label{clip_results}

\begin{figure}[ht]
    \centering
    \includegraphics[width=0.48\textwidth]{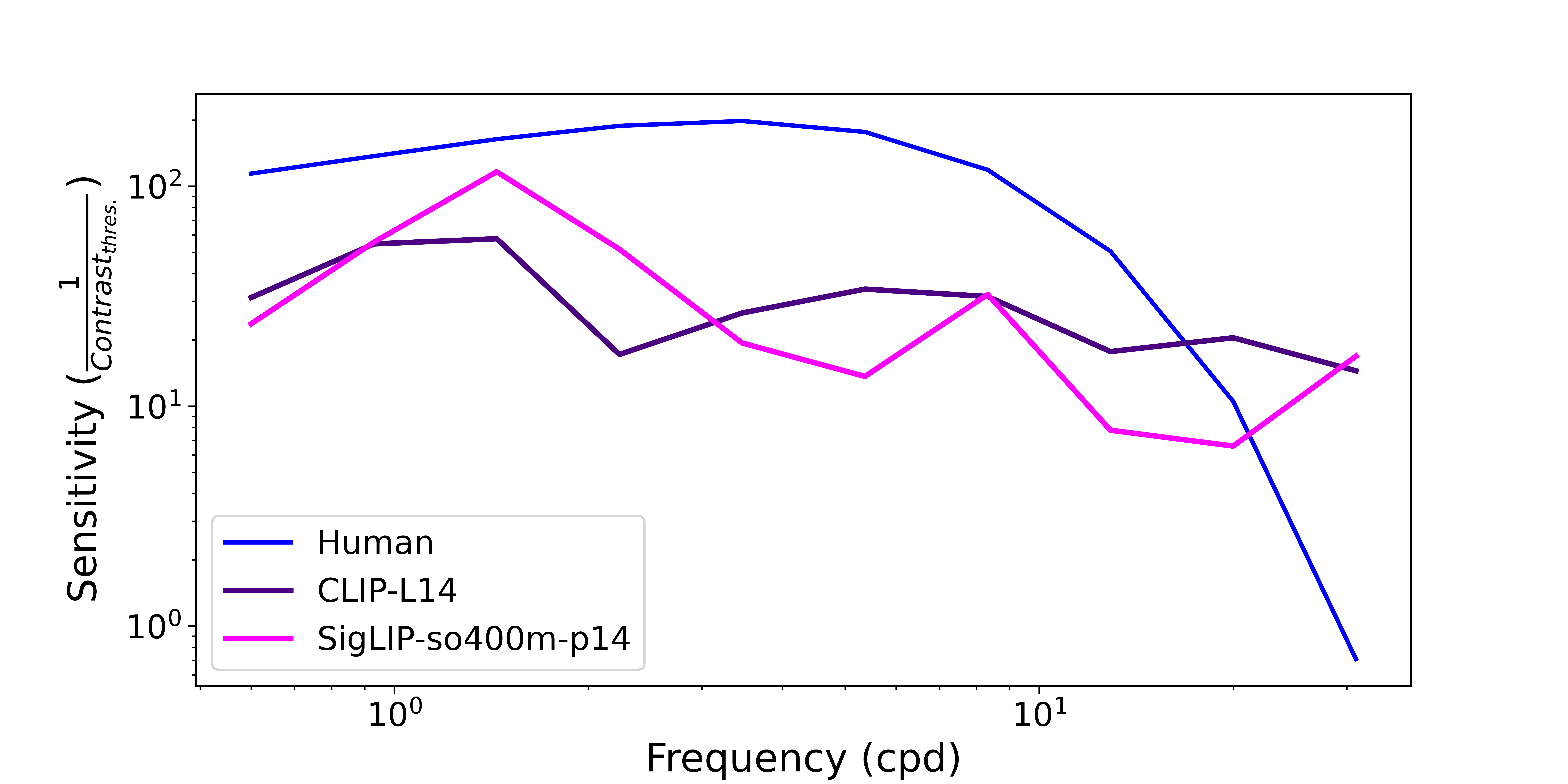}
    \caption{\textbf{Average contrast sensitivity functions (CSFs) for the two contrastive-trained tested models:} Each curve represents the average CSF across 20 prompt variants per spatial frequency. Note that we used log-log scales for better visualization. The human CSF (of the Standard Spatial Observer~\cite{Watson02}) is included for reference as the dark blue line.}
    \label{fig_csf_results_contrastive}
\end{figure}

Figure \ref{fig_csf_results_contrastive} shows the CSF obtained for the two contrastive-trained models analyzed, CLIP and SigLIP. Focusing on the results of these two models, they show a much flatter and higher CSF, indicating that they are more sensitive than the analyzed MLLMs.

This extension from our method to contrastive models such as CLIP and SigLIP, shows that CSFs can also be estimated via similarity-based readouts. This confirms the generality of our approach and highlights structural differences in perceptual behaviour across model families.

\end{document}